\documentclass[conference]{IEEEtran}
\IEEEoverridecommandlockouts
\usepackage{cite}
\usepackage{lipsum} 
\usepackage{amsmath,amssymb,amsfonts}
\usepackage[hidelinks]{hyperref}
\usepackage{algorithm}
\usepackage{array}
\usepackage{algpseudocode}
\usepackage{graphicx}
\usepackage{textcomp}
\usepackage{xcolor}
\usepackage{physics}
\usepackage{multirow}
\usepackage{subcaption}
\usepackage{mathtools}

\usepackage[left=1.62cm,right=1.62cm,top=1.9cm]{geometry}
\def\BibTeX{{\rm B\kern-.05em{\sc i\kern-.025em b}\kern-.08em
    T\kern-.1667em\lower.7ex\hbox{E}\kern-.125emX}}
\newcolumntype{M}[1]{>{\centering\arraybackslash}m{#1}}

\begin{document}

\title{CLIP Unreasonable Potential in Single-Shot Face Recognition}

\author{
    \IEEEauthorblockN{Nhan T. Luu}
    \IEEEauthorblockA{
        \textit{Department of Computer Science and Engineering}, \textit{The University of Aizu}, \\
        Aizuwakamatsu, Japan \\
        ltnhan0902@gmail.com \\
    }
}
\maketitle

\begin{abstract}

Face recognition is a core task in computer vision, designed to identify and authenticate individuals by analyzing facial patterns and features. This field intersects with artificial intelligence, image processing, and machine learning, with applications in security, authentication, and personalization. Traditional approaches in facial recognition focus on capturing facial features like the eyes, nose, and mouth and matching these against a database to verify identities. However, challenges such as high false-positive rates have persisted, often due to the similarity among individuals' facial features. Recently, Contrastive Language-Image Pretraining (CLIP), a model developed by OpenAI, has shown promising advancements by linking natural language processing with vision tasks, allowing it to generalize across modalities. Using CLIP’s vision-language correspondence and single-shot finetuning, the model can achieve lower false-positive rates upon deployment without the need of mass facial features extraction. This integration demonstrating CLIP’s potential to address persistent issues in face recognition model performance without complicating our training paradigm.
\end{abstract}


\section{Introduction}

Face recognition \cite{wang2022survey, kortli2020face, wang2021deep} is a pivotal task within the realm of computer vision, wherein algorithms are designed to identify and authenticate individuals by analyzing and comparing patterns in facial features. It's a multifaceted field that intersects with various disciplines like artificial intelligence, image processing and machine learning.

At its core, facial recognition involves capturing facial images and videos, extracting unique characteristics such as the arrangement of eyes, nose, and mouth, and then matching these features against a database of known faces to make identifications or verifications. This technology has found widespread applications across diverse sectors, ranging from security and surveillance to authentication and personalization.

CLIP\cite{radford2021learning}, or Contrastive Language-Image Pretraining, is a ground breaking model developed by OpenAI that transcends traditional boundaries between natural language processing and computer vision. Unlike conventional models that specialize in either text or image modeling, CLIP learns to understand both modalities simultaneously. This means it is possible that CLIP can perform a wide range of tasks across different domains, including those related to facial recognition.

One of the big problem distinct facial recognition with other classification task is subjectivity to false-positive result, misunderstanding someone face with others. Upon testing, we realize that by using the vision-language correspondence from CLIP features, model only require a single shot finetuning while reducing false-positive results significantly without any state-of-the-art methods for extracting facial features from massive datasets.

\begin{figure*}[htbp]
\centerline{\includegraphics[scale=0.3]{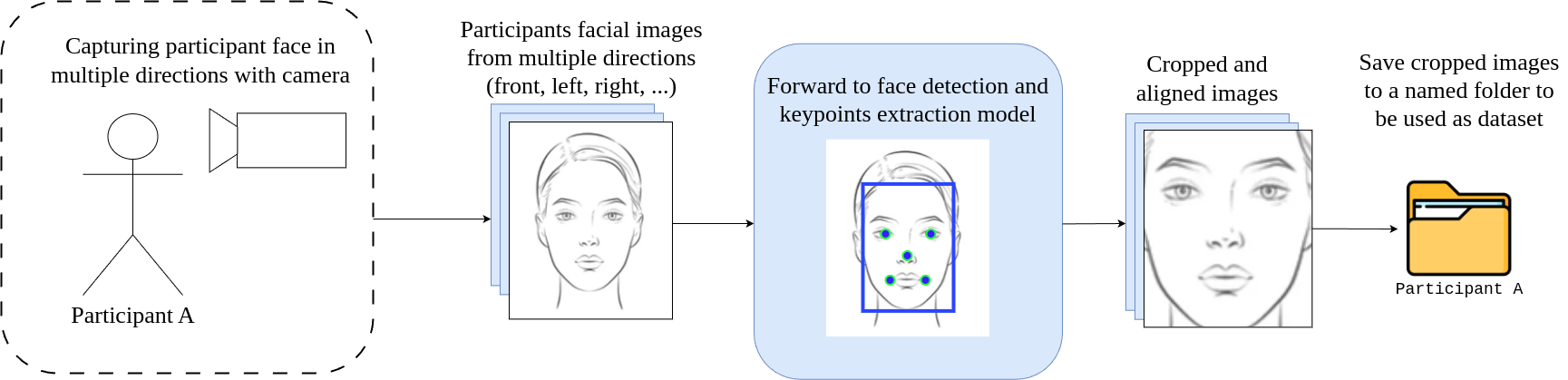}}
\caption{A diagram illustrate our dataset acquisition and preprocessing method.}
\label{fig:dataset_gather}
\end{figure*}

\section{Background}
\subsection{Face recognition}

Face recognition has emerged as a powerful tool in computer vision, enabling the identification or verification of individuals based on their facial features. This technology has applications in various fields such as security, law enforcement, and personal device authentication. Early approaches focused on geometrical models and eigenface techniques \cite{turk1991face}. With the advent of machine learning, methods such as Local Binary Patterns (LBP) \cite{ahonen2006face} and Fisherfaces \cite{belhumeur1997eigenfaces} improved robustness under various lighting conditions and facial expressions. 

The rise of deep learning has further revolutionized the field, with convolutional neural networks (CNNs) becoming the backbone of face recognition systems. Groundbreaking models like DeepFace \cite{taigman2014deepface} and FaceNet \cite{schroff2015facenet} introduced face embeddings that provided remarkable accuracy and efficiency, even in large-scale applications. More recent advancements leverage deep residual networks and novel loss functions to achieve even higher accuracy \cite{parkhi2015deep, wang2018cosface}. These advancements underscore the rapid development of face recognition and its growing importance across technology sectors.

\subsection{CLIP and its applications}

CLIP\cite{radford2021learning} model has made significant strides in connecting vision and language modalities, enabling zero-shot capabilities across various computer vision tasks \cite{zhou2023zegclip, wang2023clipn, sanghi2022clip}. CLIP\cite{radford2021learning}  leverages a contrastive learning framework, aligning text descriptions and images in a shared embedding space. This alignment allows CLIP\cite{radford2021learning} to generalize across tasks without the need for task-specific training. CLIP\cite{radford2021learning} has shown versatility in object detection\cite{vidit2023clip, gu2021open}, image generation\cite{ramesh2022hierarchical, patashnik2021styleclip}, and scene understanding \cite{zhou2022learning}. 

CLIP\cite{radford2021learning} has also been used to improve the quality and relevance of images produced by decoder-like generative models. Its capability to evaluate semantic similarity has enabled integration with autoregressive model and diffusion model such as unCLIP\cite{ramesh2022hierarchical} as well as GAN in the case of StyleCLIP\cite{patashnik2021styleclip}, where CLIP-guided latent manipulation achieves fine-grained control over generated images. Recent research has shown that CLIP-guided methods achieve better alignment with desired textual descriptions, as seen in text-to-image generation\cite{ramesh2022hierarchical, tao2023galip, wang2022clip} and image captioning\cite{barraco2022unreasonable, mokady2021clipcap} tasks. Aside from that, there are also some researches divert into improving CLIP performance by alternating its architecture, such as MaPLe\cite{luu2023blind, khattak2023maple}, CoOp\cite{zhou2022conditional} and Co-CoOp\cite{zhou2022learning}.

In face recognition, CLIP's multimodal capabilities provide unique advantages. While traditional face recognition relies on supervised learning with large labeled datasets \cite{schroff2015facenet, wang2018cosface}, recent studies have explored using CLIP to develop face recognition systems by leveraging its generalized embeddings\cite{bhat2023face, shen2023clip}. Still, CLIP face recognition potentials are still remain underexplored.

\section{Data Acquisition and Processing}

In this research, we used an image dataset consists of high resolution images of volunteers, contains faces sorted by individual names, ensuring a clear and organized structure for face recognition experiments. Images were captured using a 3 megapixels camera within a 2 meters range to obtain various perspectives of each participant, including views from above, below, left, right, and the front (as shown in Figure \ref{fig:dataset_gather}). These multiple angles simulate real-world scenarios where faces may appear in diverse orientations. Each participant contributed around 30 images, resulting in a dataset of 300 images covering 10 distinct individuals.

Due to the sensitive nature of this dataset (comprising unique facial features of each participant), only the dataset gathering methodology will be publicized to safeguard the privacy of the individuals involved.

Once the images were collected, we processed them using the SCRFD\cite{guo2021sample} face detection model. This model enabled us to detect and crop out the face region from each image accurately. For further alignment and standardization, we used the model to extract essential facial keypoints, specifically the positions of the eyes, nose, and mouth. These keypoints facilitated the vertical alignment of faces across different images, ensuring consistency in the data representation. The cropped and aligned face images were then quality assured manually by a person before being saved into a separate folder structure, labeled with each participant's name, allowing organized and efficient access for subsequent experimentation and analysis.

\section{Experiment design}

\subsection{Model training}

\begin{figure*}[htbp]
\centerline{\includegraphics[scale=0.27]{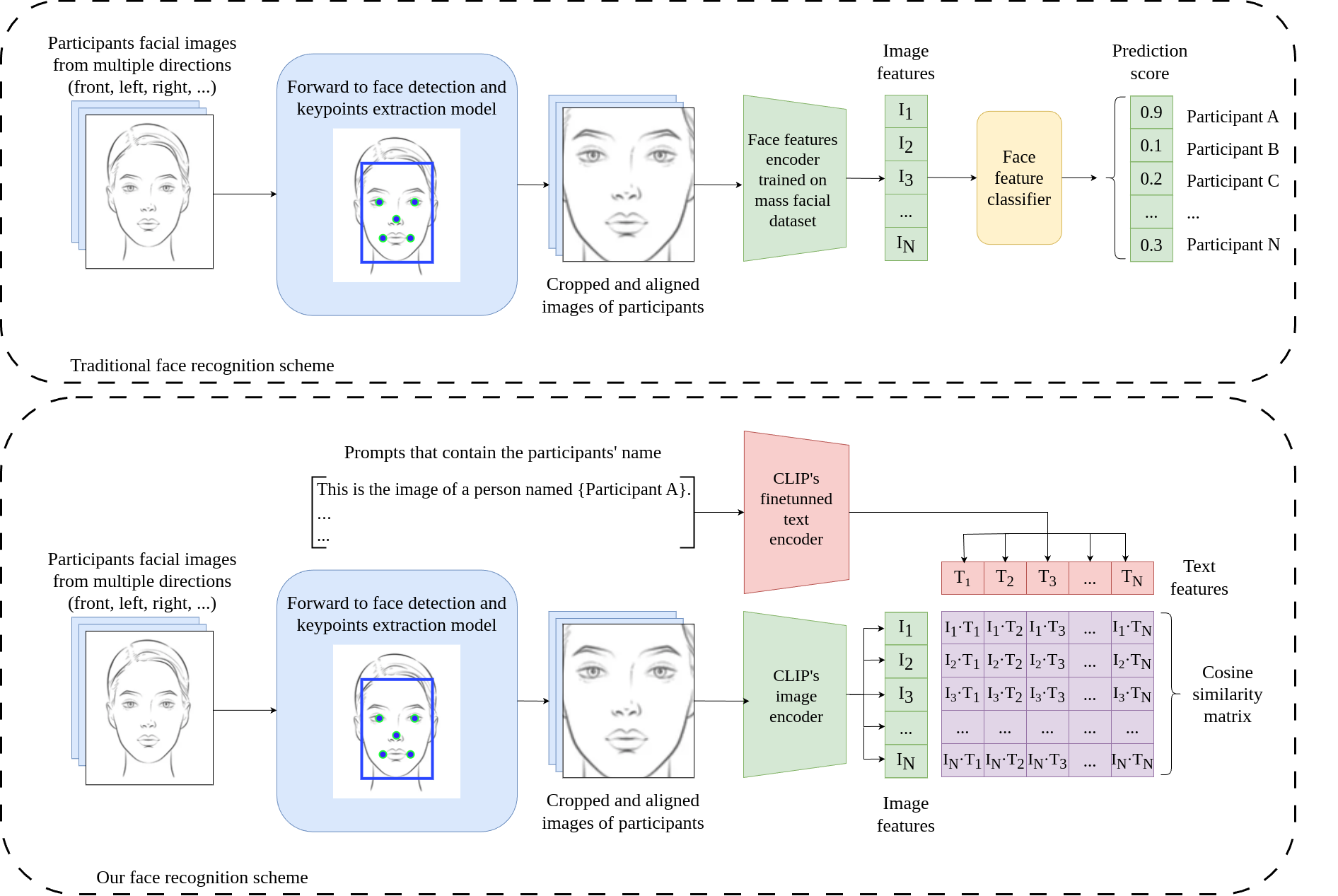}}
\caption{A graph comparing traditional face recognition pipeline and our method using single-shot finetunned CLIP model.}
\label{fig:inference_scheme}
\end{figure*}

After completing the dataset creation and preprocessing stages, we designed a face recognition pipeline that diverges from traditional approaches by replacing the standard facial feature encoder and classifier with CLIP\cite{radford2021learning} model, specifically the CLIP-RN50 variant (as shown in Figure \ref{fig:inference_scheme}). 

In this setup, we treated face recognition as an image classification task, using a single-shot finetuning method on the processed images of participants. Instead of tuning the entire model, we froze the image encoder parameters to leverage CLIP’s pretrained vision features and performed backpropagation only on the text encoder. Each image was associated with a text prompt formatted as "This is the image of a person named ..." which allows the CLIP model to align specific participant identities with their visual features.

Initially, approaching face recognition as an image classification problem may seem counterintuitive. Conventional approaches that treat face recognition this way often achieve high accuracy in training and validation phases. However, these methods typically suffer from high false-positive rates in real-world applications due to factors such as environmental noise and limited data variation. Unlike these traditional models, CLIP demonstrates robustness against false positives when deployed in real-world settings, likely due to its multimodal design and pretraining on diverse, real-world data.

Learning from previous researches related to CLIP finetuning\cite{wang2023exploring, zhang2023blind}, the model was then single-shot trained using the AdamW optimizer\cite{loshchilov2017decoupled} with a weight decay regularization parameter set to $10^{-3}$. We set an initial learning rate of $5 \times 10^{-6}$, utilizing a cosine annealing schedule\cite{loshchilovstochastic} to adjust the learning rate dynamically throughout training. Cross-Entropy Loss with Softmax function and mean reduction $L(x, y)$\cite{paszke2019pytorch} across $N$ mini-batch dimension was also widely employed as the loss function:
\begin{equation}
    \begin{aligned}
        & L(x, y) = \frac{\sum_{n=1}^{N}l_{n}}{N}, \\
        & l_{n} = -\sum_{c=1}^{C}w_{c}\log\frac{\exp{x_{n,c}}}{\sum_{i=1}^{C}\exp{x_{n,i}}}y_{n,c}
    \end{aligned}
\end{equation}
where $l_{n}$ is loss of each mini-batch, $x$ is the input, $y$ is the target, $w$ is the weight, $C$ is the total amount of class where $i \in C$ and $c$ is the mini-classes.

The training process was carried out for our dataset where 80\% of the images were selected with deterministic random seeding to form the training set, while the remaining 20\% were allocated for testing. Finetunning and inference are performed with a resized minibatch size of 16 224x224 pixels images on a single NVIDIA GeForce RTX 3090 GPU with 24GB of VRAM. This configuration optimized the model's performance efficiently while leveraging the computational power available.

\begin{table*}[htbp]
    \centering
    \begin{tabular}{|M{4cm}*{6}{|M{3cm}}|}
    \hline
    \textbf{Models} & \textbf{Training Accuracy} (in \%, higher is better)& \textbf{Deployment Accuracy} (in \%, higher is better)& \textbf{FPR} (in \%, lower is better) & \textbf{FNR} (in \%, lower is better)\\
    \hline
    VGG-Face (o)\cite{parkhi2015deep} & \textbf{96.85} & 8.33 & 90.00 & 100.00\\
    \hline
    VGG-Face (c)\cite{parkhi2015deep} & 92.72 & 0.00 & 100.00 & 100.00\\
    \hline
    Arcface (o)\cite{deng2019arcface} & 95.26 & 16.67 & 90.00 & 50.00\\
    \hline
    Arcface (c)\cite{deng2019arcface} & 94.13 & 8.33 & 90.00 & 100.00\\
    \hline
    CLIP RN-50 (Ours)\cite{radford2021learning} & 92.18 & \textbf{75.00} & \textbf{20.00} & \textbf{50.00}\\
    \hline
    \end{tabular}
    \caption{Comparison of evaluation metrics among tested models, best performance are highlighted with bold. Models denoted with "o" is trained from scratch using original settings mentioned in their paper and "c" is trained from scratch using normal image classification settings.}
    \label{tab:acc_compare}
\end{table*}

\subsection{Deployment evaluation}

To evaluate the false positive rate of our face recognition system in real-world scenarios, we conducted a deployment test of the finetuned CLIP\cite{radford2021learning} model on the same device and camera setup used during the training and data extraction phases. This consistency ensured that any environmental variables or device-specific nuances affecting model performance would closely match those encountered during initial data collection.

In this test, we instructed each of the 10 volunteer participants included in the training dataset, as well as 2 additional participants who were not in the dataset, to stand individually in front of the camera for a duration of 5 seconds. Each participant took turns, ensuring no overlap in presence before moving to the next individual. This process allowed us to observe the model’s real-time response to both known and unknown faces, assessing its ability to correctly identify or reject individuals.

Following these observations, we perform evaluation the model's deployment accuracy, false positive rate (FPR) and false negative rate (FNR) as:
\begin{equation}
    Accuracy = \frac{TP + TN}{TP + TN + FP + FN}
\end{equation}
\begin{equation}
    FPR = \frac{FP}{FP + TN}
\end{equation}
\begin{equation}
    FNR = \frac{FN}{FN + TP}
\end{equation}
where we abbreviated the values as:
\begin{itemize}
  \item \textit{TP}: True positives 
  \item \textit{TN}: True negatives
  \item \textit{FP}: False positives 
  \item \textit{FN}: False negatives
\end{itemize}
and all predictions with a confidence value lower than 80\% are considered unrecognized (classified as either false negatives or true negatives).

\section{Results}

In addition to the finetuned CLIP\cite{radford2021learning} model, we employed several well-known face recognition models, including VGG-Face \cite{parkhi2015deep} and ArcFace \cite{deng2019arcface}, for comparison under multiple settings. These settings included a traditional image recognition configuration (denoted "c") and the specific configurations originally described in each respective paper (denoted "o"), as detailed in Table \ref{tab:acc_compare}.

To ensure consistency in training and comparison, all additional models were trained using the Adam optimizer\cite{kingma2014adam}, with a learning rate of \( lr = 1 \times 10^{-3} \) and parameters \( \beta = (0.9, 0.999) \). Training was conducted over 100 epochs with identical batch sizes, resolution settings, and device configurations to CLIP\cite{radford2021learning} to maintain comparability across models.

The results show that, while the CLIP\cite{radford2021learning} experienced lower performance during the training phase in comparison to the VGG-Face\cite{parkhi2015deep} and ArcFace\cite{deng2019arcface} variants, it demonstrated significantly improved performance during deployment inference. Notably, the finetuned CLIP\cite{radford2021learning} achieved a markedly lower FPR and a reduced FNR compared to the other models. This enhanced performance was achieved even though CLIP\cite{radford2021learning} was fine-tuned with a single-shot approach as an image recognition model, without employing any advanced training techniques or traditional feature extraction methods used in face recognition tasks.

Although we observed a significant degradation in performance across multiple variants of VGG-Face \cite{parkhi2015deep} and ArcFace \cite{deng2019arcface}, this outcome was somewhat expected, as these models were originally designed for large-scale facial feature extraction. When applied to an image recognition scenario or when only a limited set of facial features are used, their performance diminishes drastically, highlighting the models' dependency on detailed facial feature analysis for optimal results.

The results highlight CLIP's robustness and effectiveness in real-world deployment scenarios, making it a competitive choice for face recognition tasks with minimal training adjustments.

\section{Discussion}
\subsection{Prompting choices}
While there has been considerable research on prompting optimization for CLIP in various computer vision tasks, there is a noticeable lack of studies addressing this specifically within the context of face recognition. As part of our experiments, we explored the effect of different prompt formulations on the CLIP model's performance in face recognition tasks.

In addition to the standard prompt "This is the image of a person named ...", we tested other variations, such as simply stating "This is the image of ..." and directly feeding the name of the person to the text encoder without any additional context. Despite these variations, the fine-tuning results exhibited only minimal fluctuations in performance, with a difference of approximately 1\%. This suggests that, in the context of face recognition, the choice of prompt and the inclusion of additional context may not significantly impact the model's performance.

\subsection{Distinctive features between faces}

While CLIP achieves a lower FPR compared to the majority of tested models, its performance remains far from perfect. As observed from the results, although both the FPR and FNR are reduced relative to other tested models, this level of performance is still inadequate for security-sensitive applications of face recognition. 

During testing, we noticed a relatively large cosine similarity between face features produced by CLIP's image encoder (approximately 80\%). We believe this high similarity contributes significantly to CLIP's suboptimal performance when applied as part of a face recognition pipeline. To address this limitation, we propose that future researches could focus on improving the model by training with a triplet sampling method, as utilized in FaceNet\cite{schroff2015facenet}, or by incorporating ArcFace loss\cite{deng2019arcface} during the training process.

\subsection{Training gradient}
During the hypothesis formulation and experiment design phase, we initially planned to test on a significantly larger number of classes, typical of traditional large-scale face verification problems where the number of classes greatly exceeds the number of images per class. This would have allowed us to explore the scalability and robustness of CLIP in more complex, real-world face recognition tasks.

However, during testing, we encountered a significant limitation. We realized that to handle such a large number of classes, a substantial amount of GPU memory would be required, particularly because the gradients of CLIP's text encoder grow very large as the number of classes increases. Given our limited computing resources, we were unable to scale the problem as initially planned and were forced to downscale the experiment to just 10 classes (equivalent to 10 persons). This reduction in scale highlights a potential challenge when attempting to apply CLIP to large-scale image databases, as the computational demands may become prohibitive without access to high-performance hardware. This issue may need to be addressed in future work to enable the use of CLIP in more expansive and computationally intensive face recognition scenarios.

\section{Conclusion}
In this paper, we explored the application of Contrastive Language-Image Pretraining (CLIP) for face recognition, highlighting its potential to address key challenges in the field. Traditional face recognition models typically rely on extracting and matching detailed facial features, but they often struggle with high false-positive rates due to the inherent similarity among individuals' facial structures. By leveraging CLIP's vision-language correspondence and employing a single-shot finetuning approach, we demonstrated that the model can achieve significantly lower false-positive rates during deployment while eliminating the need for complex facial feature extraction techniques.

Our findings indicate that CLIP, with its ability to generalize across modalities, offers a promising solution to persistent issues in face recognition systems. The model’s robustness in real-world deployments, without the need for extensive training data or traditional feature extraction methods, positions it as a powerful alternative to conventional face recognition models. Future work may focus on scaling CLIP to larger datasets and further optimizing its performance for diverse and dynamic real-world applications. Overall, this research contributes to advancing face recognition technology, showcasing CLIP’s potential for more efficient and accurate facial authentication systems.

\section*{Acknowledgement}
Special thanks to Can Tho University Software Center R\&D Department's researchers for their insights as well as assistance in dataset gathering process.

\bibliographystyle{IEEEtran}
\bibliography{refs}

\end{document}